\documentclass{bmvc2k}

\usepackage{times}
\usepackage{xcolor}
\usepackage{soul}
\usepackage{float}
\usepackage[utf8]{inputenc}
\usepackage{booktabs}
\usepackage{amssymb}
\usepackage{colortbl}
\usepackage{graphicx}
\usepackage{amsmath}
\usepackage{makecell}
\usepackage{multirow}
\usepackage{tabu}
\usepackage{array}
\usepackage{bbm}
\usepackage{hyperref}

\graphicspath{ {images/} }

\title{Pose Flow: Efficient Online Pose Tracking}

\addauthor{Yuliang Xiu}{yuliangxiu@sjtu.edu.cn}{1}
\addauthor{Jiefeng Li}{ljf_likit@sjtu.edu.cn}{1}
\addauthor{Haoyu Wang}{why2011btv@sjtu.edu.cn}{1}
\addauthor{Yinghong Fang}{yhfang@sjtu.edu.cn}{1}
\addauthor{Cewu Lu}{lu-cw@cs.sjtu.edu.cn}{1}

\addinstitution{
 Machine Vision and Intelligence Group\\
 Shanghai Jiao Tong University\\
 Shanghai, China
}

\runninghead{Student, Prof, Collaborator}{BMVC Author Guidelines}

\def\etal{\emph{et al}\bmvaOneDot}

\begin{document}

\maketitle

\begin{abstract}
Multi-person articulated pose tracking in unconstrained videos is an important while challenging problem. In this paper, going along the road of top-down approaches, we propose a decent and efficient pose tracker based on pose flows. First, we design an online optimization framework to build the association of cross-frame poses and form pose flows (PF-Builder). Second, a novel pose flow non-maximum suppression (PF-NMS) is designed to robustly reduce redundant pose flows and re-link temporal disjoint ones. Extensive experiments show that our method significantly outperforms best reported results on two standard Pose Tracking datasets (~\cite{iqbal2017posetrack} and~\cite{girdhar2017detect}) by \textbf{13 mAP 25 MOTA} and \textbf{6 mAP 3 MOTA} respectively. Moreover, in the case of working on detected poses in individual frames, the extra computation of  pose tracker is very minor, guaranteeing online \textbf{10FPS} tracking. Our source codes are made \textbf{publicly available}\footnote{\href{url}{https://github.com/YuliangXiu/PoseFlow}}.
\end{abstract}

\section{Introduction}
\label{sec:intro}

Motivated by its extensive applications in human behavior understanding and scene analysis, human pose estimation has witnessed a significant boom in recent years. Mainstream research fields have advanced from pose estimation of single pre-located person \cite{newell2016stacked,chu2017multi} to multi-person pose estimation in complex and unconstrained scenes \cite{cao2017realtime,fang2017rmpe}. Beyond static human keypoints in individual images, pose estimation in videos has also emerged as a prominent topic  \cite{song2017thin,zhang2015human}. Furthermore, human pose trajectory extracted from the entire video is a high-level human behavior representation \cite{wang2013action,wang2015action}, naturally providing us with a powerful tool to handle a series of visual understanding tasks, such as Action Recognition  \cite{cheron2015p}, Person Re-identification \cite{su2017pose,zheng2017pose}, Human-Object Interaction  \cite{gkioxari2017detecting} and numerous downstream practical applications, e.g., video surveillance and sports video analysis.

To this end, multi-person pose tracking methods are developed, whose dominant approaches can be categorized into top-down \cite{girdhar2017detect} and bottom-up \cite{insafutdinov2017arttrack,iqbal2017posetrack}. Top-down methods, also known as two steps scheme, first detect human proposals in every frame, estimate  keypoints within each box independently, and then track human boxes over the entire video in terms of similarity between pairs of boxes in adjacent frames, and that is the reason why it is also referred to as Detect-and-Track method \cite{girdhar2017detect}. By contrast, bottom-up methods, also known as jointing scheme, first generate a set of joint detection candidates in every frame, construct the spatio-temporal graph, and then solve an integer linear program to partition this graph into sub-graphs that correspond to plausible human pose trajectories of each person.

\begin{figure}[!ht]
	\begin{center}
	\includegraphics[scale=0.37]{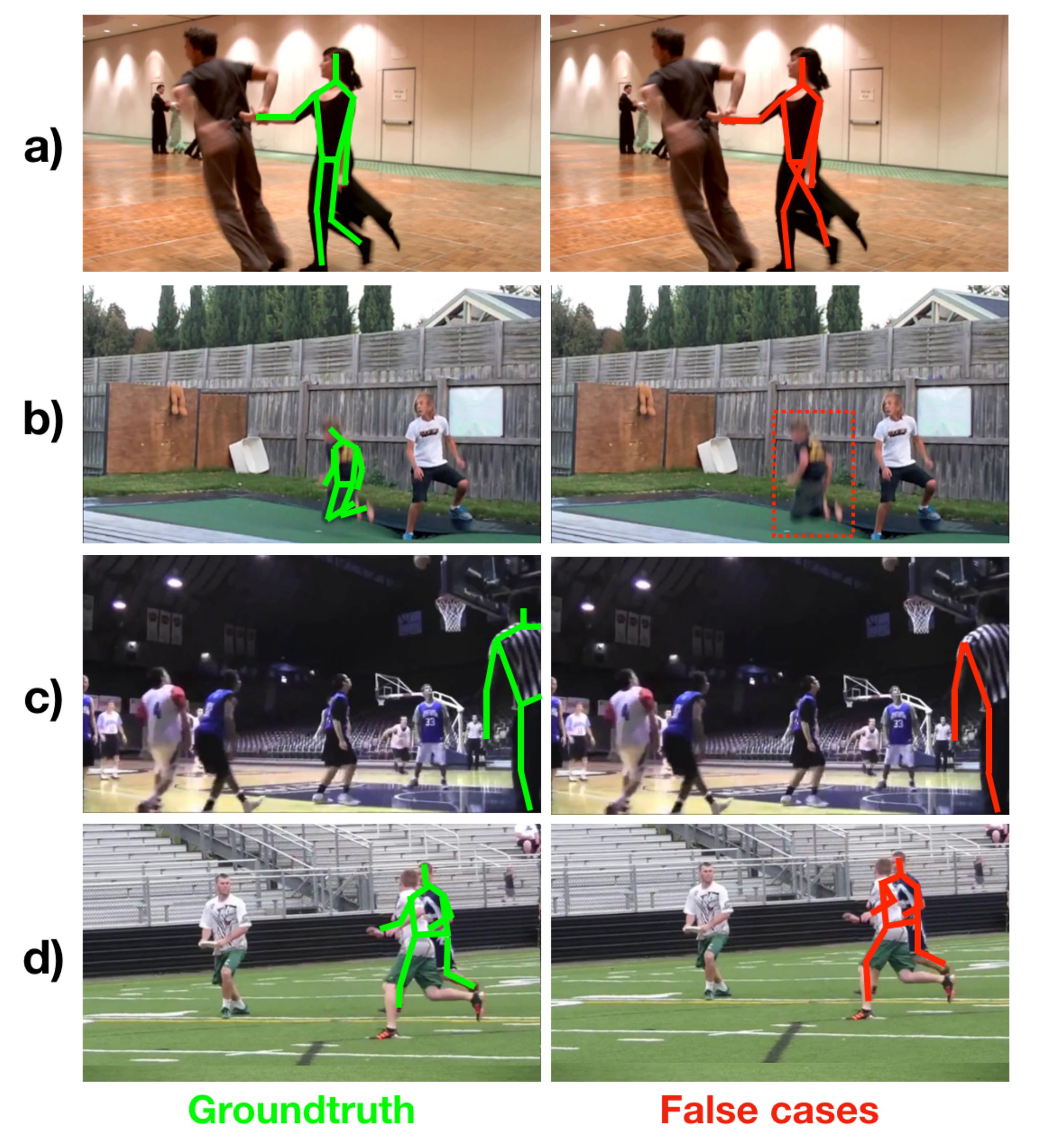}
	\caption{Failure cases of previous pose estimation methods, ground-truth in green and false cases in red. a) Ambiguous assignment. b) Missing detection. c) Human truncation. d) Human occlusion.}
	\label{Figure:falses}
	\end{center}
\end{figure}

Currently top-down methods have largely outperformed bottom-up methods both in accuracy (mAP and MOTA) and tracking speed, since bottom-up approaches lose a global pose view due to the mere utilization of second-order body parts dependence, which directly cause ambiguous assignments of keypoints, like Figure \ref{Figure:falses} a). Furthermore, joint schemes are computationally heavy and not scalable to long videos, making it unable to do online tracking. Therefore, top-down methods may be a more promising direction. Following this direction, however, there remains many challenges. As shown in Figure \ref{Figure:falses} b) c) d), due to frame degeneration (e.g. blurring due to fast motion), truncation or occlusion, pose estimation in an individual frame can be unreliable. To tackle this problem, we need to associate cross-frame detected instances to share temporal information and thus reduce uncertainty. 

\vspace{-4mm}
\vspace{-7mm}
In this paper, we propose an efficient and decent method to achieve online pose tracking. Apart from applying an improved RMPE\cite{fang2017rmpe} as pose estimator, our proposed method includes two novel techniques, namely Pose Flow Building (PF-Builder) and Pose Flow NMS(PF-NMS). First, we associate the cross-frame poses that indicate the same person. To achieve that, we iteratively construct pose flow from pose proposals within a short video clip picked by a temporal video sliding window. Instead of employing greedy match, we design an effective objective function to seek a pose flow with maximum overall confidence among potential flows. This optimization design helps to stabilize pose flows and associate discontinuous ones (due to missing detections). Second, unlike conventional schemes that apply NMS in frame-level, PF-NMS takes pose flow as a unit in NMS processing. In this way, temporal information will be fully considered in NMS process and thus stabilization can be largely improved. Our approach is general to different pose estimators and only takes minor extra computation for tracking. Given detected poses in individual frames, our method can track poses at 10 FPS. 

To verify the effectiveness of proposed framework, we conduct extensive experiments on two standard pose tracking datasets, \textbf{PoseTrack Dataset} \cite{iqbal2017posetrack} and \textbf{PoseTrack Challenge Dataset} \cite{andriluka2017posetrack}. Our proposed approach significantly outperforms the state-of-the-art method  \cite{girdhar2017detect}, achieving 58.3\% MOTA and 66.5\% mAP in PoseTrack Challenge validation set, 51.0\% MOTA and 63.0\% mAP in testset. 

\section{Related Work}

\subsection{Multi-Person Pose Estimation in Image}

In recent years, multi-person pose estimation in images has experienced large performance advancement. With respect to different pose estimation pipelines, relevant work can be grouped into graph decomposition and multi-stage techniques. Graph decomposition methods, such as DeeperCut \cite{insafutdinov2016deepercut}, re-define the multi-person pose estimation problem as a partitioning and labeling formulation and solve this graph decomposition problem by an integer linear program. These methods' performance depends largely on strong parts detector based on deep visual representations and efficient optimization strategy. However, their body parts detector always performs vulnerably because of the absence of global context and structural information. OpenPose \cite{cao2017realtime} introduces Part Affinity Fields (PAFs) to associate body parts with individuals in an image, but ambiguous assignments still occur in crowds. 

To address this limitation, multi-stage pipeline \cite{fang2017rmpe,chen2017cascaded} handles multi-person pose estimation problem by separating this task into human detection, single person pose estimation and post-processing stages. The main difference among dominant multi-stage frameworks lies in different choices of the human detector and single person pose estimator network. With the remarkable progress of object detection and single person pose estimator over the past few years, the potentials of multi-stage approaches have been greatly exploited. Now multi-stage framework has been in the epicenter of the methods above, achieving the state-of-the-art performance in almost all benchmark datasets, e.g., MSCOCO\cite{lin2014microsoft} and MPII\cite{andriluka14cvpr}.
\vspace{-2mm}
\subsection{Multi-Person Articulated Tracking in Video}
Based on the multi-person pose estimators described above, it is natural to extend them from still image to video. PoseTrack \cite{iqbal2017posetrack} and ArtTrack \cite{insafutdinov2017arttrack} in CVPR'17 primarily introduce multi-person pose tracking challenge and propose a new graph partitioning formulation, building upon 2D DeeperCut  \cite{insafutdinov2016deepercut} by extending spatial joint graph to spatio-temporal graph. Although plausible results can be guaranteed by solving minimum cost multicut problem, hand-crafted graphical models are not scalable for long clips of unseen types of scenes. It is worth noting that optimize this sophisticated IP requires tens of minutes per video, even implemented with state of the art solvers.  

Hence, another line of research tends to explore more efficient and scalable top-down method by first operating multi-person pose estimation on each frame, and then link them in terms of appearance similarity and temporal relationship between pairs of boxes. Yet some issues should be dealt with properly: 1) how to filter redundant boxes correctly with the fusion of information from adjacent frames, 2) how to produce robust pose trajectories by leveraging temporal information, 3) how to connect human boxes with the same identity meanwhile keeping away from disturbance of scale variance. 

Although one latest work, 3D Mask R-CNN\cite{girdhar2017detect}, which is designed for correcting the location of keypoints by leveraging temporal information in 3D human tubes, tries to give their solution to these problems, it do not employ pose flow as a unit. Besides, the tracker just simplify tracking problem as a maximum weight bipartite matching problem and solve it with greedy or Hungarian Algorithm. Nodes of this bipartite graph are human bounding boxes in two adjacent frames. This configuration did not take motion and pose information into account, which is essential in tracking the occasional truncated human. To address this limitation, meanwhile maintaining its efficiency, we put forward a new pose flow generator, which combines Pose Flow Builder and Pose Flow NMS. 

\begin{figure*}[!ht]
	\begin{center}
		\includegraphics[width=\linewidth]{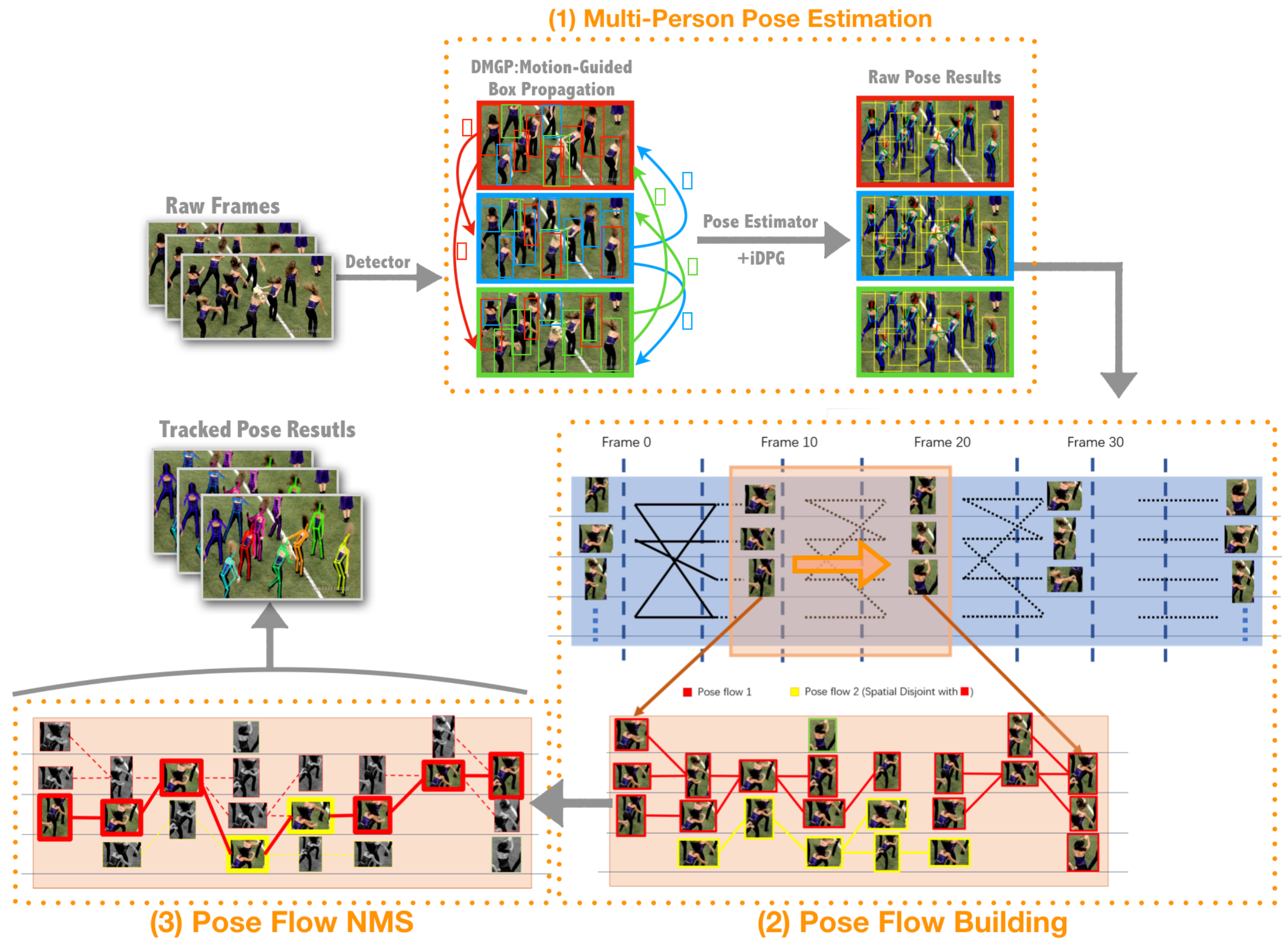}
		\caption{Overall Pipeline: 1) Pose Estimator. 2) Pose Flow Builder. 3) Pose Flow NMS. First, we estimate multi-person poses. Second, we build pose flows by maximizing overall confidence and purify them by Pose Flow NMS. Finally, reasonable multi-pose trajectories can be obtained.}
		\label{Figure:Pipeline}
	\end{center}
	\vspace{-8mm}
\end{figure*}

\section{Our Proposed Approach}
\vspace{-2mm}
In this section, we present our pose tracking pipeline. As mentioned before, pose flow means a set of pose indicating the same person instance in different frames. As Figure \ref{Figure:Pipeline} shows, our framework includes two steps: Pose Flow Building and Pose Flow NMS. First, we build pose flow by maximizing overall confidence along the temporal sequence. Second, we reduce redundant pose flows and relink disjoint pose flows by Pose Flow NMS. 
\vspace{-4mm}
\subsection{Preliminary}
\vspace{-2mm}
In this section, we introduce some basic metrics and tools that will be used in our framework.
\vspace{-6mm}
\paragraph{Intra-Frame Pose Distance}
Intra-frame Pose distance is defined to measure the similarity between two poses $P_1$ and $P_2$ in a frame. We adopt the pose distance defined in  \cite{fang2017rmpe}. We denote $p_1^n$ and $p_2^n$ as the $n^{th}$ keypoints of pose $P_1$ and $P_2$ respectively, $n \in \{1,2,...,N\}$, $N$ is keypoint number of one person, $B(p_1^n)$ is box that centers at $p_1^n$, $c_1^n$ is score of $p_1^n$. The $tanh$ function is to suppress the low score keypoints. 

The soft matching function is defined as
\begin{equation}
\begin{aligned}
    &K_{Sim}(P_1,P_2| \sigma_1) = \\
    \quad
    &\left\{
    \begin{aligned}
    &\sum_{n} tanh \frac{c_1^n}{\sigma_1} \cdot tanh \frac{c_2^n}{\sigma_1} &\text{if } p_2^n \text{ is within } B(p_1^n) \\
    &0 &otherwise \\
    \end{aligned}
    \right.
    \label{eq:soft}
\end{aligned}
\end{equation}

The spatial similarity among keypoints written as
\begin{equation}
H_{Sim}(P_1,P_2|\sigma_2) = \sum_n exp[-\frac{(p_1^n-p_2^n)^2}{\sigma_2}]
 \label{eq:spatial}
\end{equation}

The final similarity combining Eqs. \ref{eq:soft} and \ref{eq:spatial} is written as
\begin{equation}
\begin{aligned}
&d_f(P_1, P_2|\Lambda) \\
&= K_{Sim}(P_1,P_2|\sigma_1)^{-1} + \lambda H_{Sim}(P_1,P_2|\sigma_2)^{-1}
 \label{eq:dist}
\end{aligned}
\end{equation}

where $\Lambda=\{\sigma_1,\sigma_2,\lambda\}$. These parameters can be determined in a data-driven manner.
\vspace{-3mm}
\paragraph{Inter-frame Pose Distance}
Inter-frame pose distance is to measure distance between a pose $P_1$ in one frame and another pose $P_2$ in the next frame. We need to import temporal matching to measure how likely two cross-frame poses indicate the same person. Bounding boxes surrounding $p_1^n$ and $p_2^n$ are extracted and denoted as $B_1^n$ and $B_2^n$. The box size is $10\%$ person bounding box size according to the standard PCK \cite{andriluka14cvpr}. We evaluate the similarity of $B_1^n$ and $B_2^n$. Given $f_1^n$ DeepMatching feature \cite{revaud2016deepmatching} points extracted from $B_1^n$, we can find $f_2^n$ matching points in $B_2^n$. The matching percentage $\frac{f_2^n}{f_1^n}$ can indicate the similarity of $B_1^n$ and $B_2^n$. Therefore the inter-frame pose distance between $P_1$ and $P_2$ can be expressed as:
\begin{equation}
\begin{aligned}
 d_c(P_1,P_2) =  \sum_{n} \frac{f_2^n}{f_1^n}
\end{aligned}
\end{equation}

\subsection{Improved Multi-Person Pose Estimation}
We adopt RMPE \cite{fang2017rmpe} as our multi-person pose estimator, which uses Faster R-CNN\cite{ren2015faster} as the human detector and Hourglass Network with PRMs \cite{yang2017learning} as single person pose estimator. Our pipeline is ready to adopt to different human detectors and pose estimators.
\vspace{-3mm}
\paragraph{Data Augmentation} 
In video scenoria, human always come and leave video capturing region, resulting in truncation problem. To handle truncation of humans, we propose an improved deep proposal generator (iDPG) as a data augmentation scheme. iDPG aims to produce truncated human proposals using random-crop strategy during training. Specifically, we randomly crop human instance region into quarter or half person. Thus, those random-crop proposals will be used as augmented training data. We observe an improvement of RMPE when it applies to the video frames 
\vspace{-4mm}
\paragraph{Motion-Guided Box Propagation} Due to motion blur and occlusion, missing detection happens frequently during human detection phrase. This will increase person id switches (IDs$\downarrow$), like in Table \ref{Table:PoseTrack Dataset},  dramatically degrading final tracking MOTA performance. Our idea is to propagate box proposals to previous and next frames by crossing frame matching technique. That is, the box proposals triple. In this way, some missing detected proposals have high channce to be recovered and largely improve the recall (redundant boxes will be filter out by following step). The cross-frame matching technique we used is deepmatching\cite{revaud2016deepmatching}.
\vspace{-3mm}
\subsection{Pose Flow Building}
We firstly perform pose estimation for each frame. Pose flows are built by associating poses that indicate the same person across frames. The straight-forward method is to connect them by selecting closest pose in the next frame, given metric $d_c(P_1,P_2)$. However, this greedy scheme would be less effective due to recognition error and false alarm of frame-level pose detection. On the other hand, if we apply the graph-cut model in spatial and temporal domains, it will lead to heavy computation and non-online solution. Therefore, in this paper, we propose an efficient and decent method for high-quality pose flow building. We denote $P_i^j$ as the $i^{th}$ pose at $j^{th}$ frame and its candidate association set as
\vspace{-1mm}
\begin{equation}
\begin{aligned}
\mathcal{T}(P_i^j) = \{P| d_c(P,P_i^j) \leq \epsilon\}, \\
~~~s.t. P \in \Omega_{j+1}
\label{eq:set}
\end{aligned}
\end{equation}
where $\Omega_{j+1}$ is the set of pose at $(j+1)^{th}$ frame. In paper, we set $\epsilon = \frac{1}{25}$ by cross-validation. $\mathcal{T}(P_i^j)$ means possible corresponding pose set in next frame for $P_i^j$. Without loss of generality, we discuss tracking for $P_i^t$ and consider pose flow building from $t^{th}$ to $(t+T)^{th}$ frames. To optimize pose selection, we maximize the following objective function
\vspace{-2mm}
\begin{equation}
\begin{aligned}
\vspace{2mm}
F(t,T)  &= \max_{Q_t,\ldots,Q_{t+T}} \sum_{i=t}^{t+T} s(Q_i),  \\
s.t.~~&Q_0 = P_i^t ,   \\
s.t.~~&Q_i \in \mathcal{T}(Q_{i-1})
\label{eq:online}
\end{aligned}
\end{equation}
where $s(Q_i)$ is a function that outputs confidence score of $Q_i$, which is defined as
\begin{equation}
\begin{aligned}
s(Q_i) = s_{box}(Q_i) + mean(s_{pose}(Q_i)) + \max(s_{pose}(Q_i))
\end{aligned}
\end{equation}
where $s_{box}(P)$, $mean(s_{pose}(P))$ and $\max(s_{pose}(P))$ are score of human box, mean score and max score of all keypoints within this human proposal, respectively. The optimum $\{Q_t,\ldots,Q_{t+T}\}$ is our pose flow for $P_i^t$ from $t^{th}$ to $(t+T)^{th}$ frame.
\vspace{-3mm}

\paragraph{Analysis} We regard the sum of confidence scores ($\sum_{i=t}^{t+T} s(Q_i)$) as objective function. This design helps us resist many uncertainties. When a person is highly occluded or blurred, its score is quite low because the model is not confident about it. But we can still build a pose flow to compensate it, since we look at the overall confidence score of a pose flow, but instead of a single frame. Moreover, the sum of confidence score can be calculated online. That is, $F(t,T)$ can be determined by $F(t,T-1)$ and $s(Q_T)$.
\vspace{-4mm}
\paragraph{Solver} Eq. \ref{eq:online} can be solved in an online manner since it is a standard dynamic programming problem. At $(u-1)^{th}$ frame, we have $m_{u-1}$ possible poses and record $m_{u-1}$ optimum pose trajectories (with sum of scores) to reach them. At $u^{th}$ frame, we compute the optimum paths to $m_{u}$ possible poses based on previous $m_{u-1}$ optimum pose trajectories. Accordingly, $m_u$ trajectories are updated. $F(u)$ is the sum of scores of best pose trajectories. 

\subsubsection{Stop Criterion and Confidence Unification} We process video frame-by-frame with Eq. \ref{eq:online} until it meets a stop criterion. Our criterion doesn't simply check confidence score in a single frame but looks at more frames to resist sudden occlusion and frame degeneration (e.g. motion blur). Therefore, a pose flow stops at $u$ when $F(t,u+r) - F(t,u) < \gamma$, where $\gamma$ is determined by cross-validation. It means the sum of scores within the following $r$ frames is very small. Only in this way, we can make sure a pose flow really stops. In our paper, we set $r = 3$. After a pose flow stops, all keypoint confidence are updated by average confidence scores. We believe pose flow should be the basic block and should use single confidence value to represent it. This process is referred to as confidence unification.
\vspace{-3mm}
\subsection{Pose Flow NMS}
We hope our NMS can be performed in the spatio-temporal domain instead of individual frame processing. That is, we take poses in a pose flow as a unit in NMS processing, reducing errors by both spatial and temporal information. The key step is to determine the distance between two pose flows that indicate the same person.

\vspace{-4mm}
\paragraph{Pose Flow Distance}
Given two pose flows $\mathcal{Y}_a$ and $\mathcal{Y}_b$, we can extract their temporal overlapping sub-flows. The sub-flows are denoted as $\{P_a^1,\ldots, P_a^N\}$ and $\{P_b^1,\ldots,P_b^N\}$, where $N$ is the number of temporal overlapping frames. That is, $P_a^i$ and $P_b^i$ are two poses in the same frame. The distance between $\mathcal{Y}_a$ and $\mathcal{Y}_b$ can be calculated as,
\begin{equation}
\begin{split}
d_{PF}(\mathcal{Y}_a, \mathcal{Y}_b) = median[\{d_f(P_a^1, P_b^1),\ldots,d_f(P_a^N, P_b^N)\}]
\end{split}
\end{equation}
where $d_f(\cdot)$ is the intra-frame pose distance defined in Eq. \ref{eq:dist}. The median metric can be more robust towards outliers, such as miss-detection due to occlusion and motion blur.
\vspace{-3mm}

\begin{figure*}[!ht]
\centering
\includegraphics[scale=0.43]{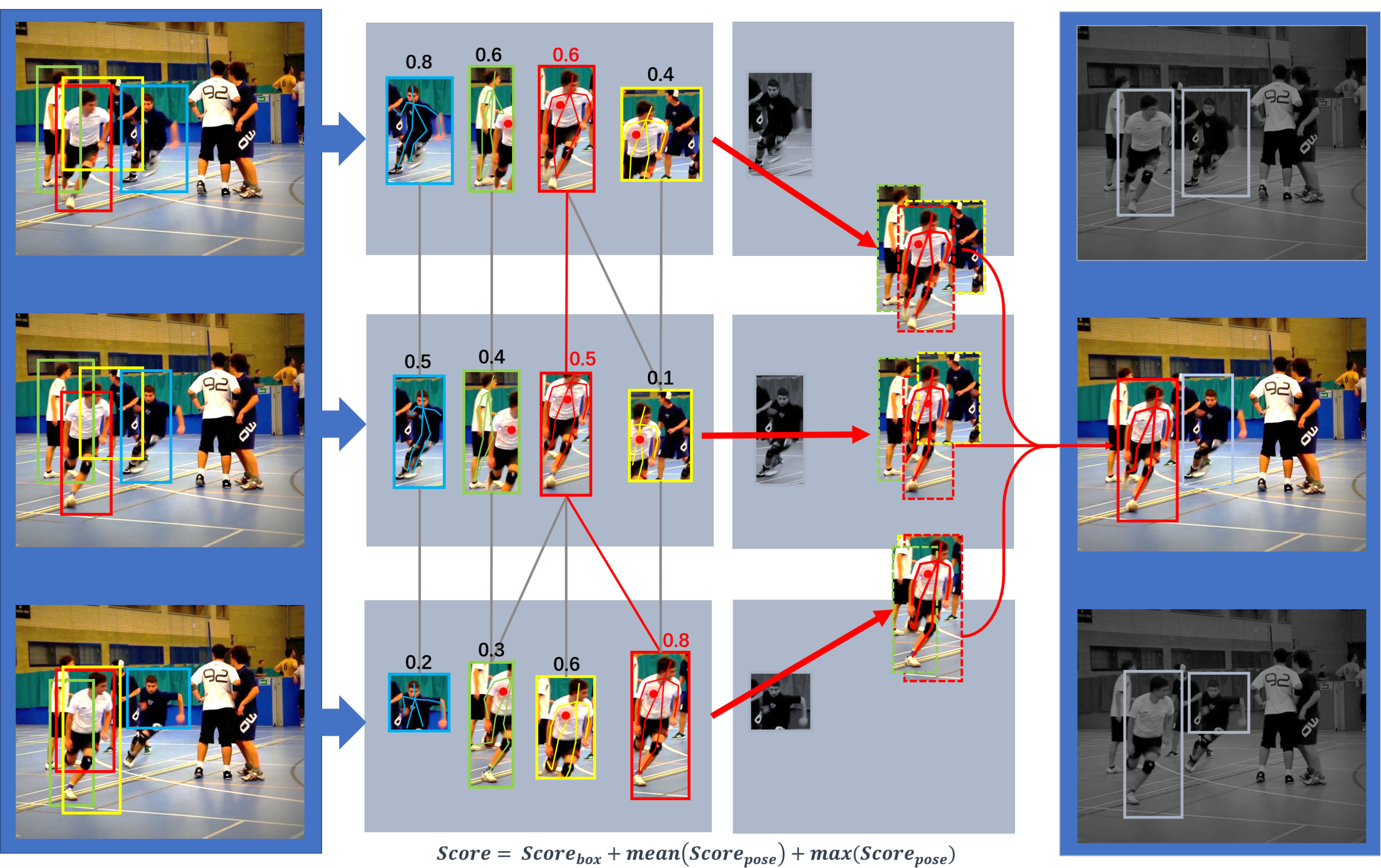} 
\caption{Pose Flow Merging}
\label{fig:nms}
\vspace{-2mm}
\end{figure*}

\paragraph{Pose Flow Merging}
Given $d_{PF}(\cdot)$, we can perform NMS scheme as convection pipeline. First, the pose flow with the maximum confidence score (after confidence unification) is selected as reference pose flow. Making use of $d_{PF}(\cdot)$, we group pose flows closed to reference pose flow. Thus, pose flows in the group will be merged into a more robust pose flow representing the group. This new pose flow 
(pose flow NMS result) is called representative pose flow. The 2D coordinate of $i^{th}$ keypoint $\textbf{x}_{t,i}$ and confidence score $s_{t,i}$ of representative pose flow in $t^{th}$ frame are computed by
\vspace{-2mm}
\begin{equation}
\begin{aligned}
 \hat{\textbf{x}}_{t,i} = \frac{\sum_{j} s_{t,i}^j \textbf{x}_{t,i}^j}{\sum s_{t,i}^j} ~~~and~~~~
 \hat{s}_{t,i} = \frac{\sum_{j} s_{t,i}^j}{\sum \mathbbm{1}(s_{t,i}^j) } \\
 \label{eq:update}
\end{aligned}
\vspace{-6mm}
\end{equation}
where $\textbf{x}_{t,i}^j$ and $s_{t,i}^j$ are the 2D coordinate and confidence score of $i^{th}$ keypoint in $j^{th}$ pose flow in the group in $t^{th}$ frame. If $j^{th}$ pose flow does not have any pose at $t^{th}$ frame, we set $s_{t,i}^j = 0$. In Eq. \ref{eq:update}, $\mathbbm{1}(s_{t,i}^j)$ outputs $1$, if input is non-zero, otherwise it outputs $0$. This merging step not only can reduce redunant pose flow, but also re-link some disjoint pose flows into a longer and completed pose flow. Details of cross-frame pose merging (keypoint-level) can be refered to Figure \ref{fig:nms}.

We redo this process until all pose flows are processed. This process is computed in sliding temporal window (the window length is $L=20$ in our paper). Therefore, it is an online process. The whole pipeline shows in Figure \ref{Figure:Pipeline}.

\vspace{-3mm}
\section{Experiments and Results}

\subsection{Evaluation and Datasets}
For comparison with both state-of-the-art top-down and bottom-up approaches, we evaluate our framework on \textbf{PoseTrack} and \textbf{PoseTrack Challenge} dataset separately. PoseTrack Dataset was introduced in  \cite{iqbal2017posetrack}, which is used to evaluate the spatio-temporal graph-cut method. Labeled frames in this dataset come from consecutive unlabeled adjacent frames of MPII Multi-Person Pose dataset\cite{andriluka14cvpr}. These selected videos contain multiple persons and cover a wide variety of activities of complex cases, such as scale variation, body truncation, severe occlusion and motion blur. For a fair comparison, we train improved RMPE on 30 training videos and test it on the rest 30 videos like PoseTrack \cite{iqbal2017posetrack} did. Table \ref{Table:PoseTrack Dataset} presents tracking results in PoseTrack dataset, and pose estimation results in Table  \ref{Table:mAP}. It shows that our method outperforms best reported graph-cut approach by \textbf{13.5 mAP} and \textbf{25.4 MOTA}.
\vspace{-1mm}
\begin{table*}[!ht]
\centering
\scalebox{0.75}{
\begin{tabular}{ccccccc>{\columncolor[gray]{.8}}cc}
\toprule
\textbf{Method}&\textbf{Rcll$\uparrow$}&\textbf{Prcn$\uparrow$}&\textbf{MT$\uparrow$}&\textbf{ML$\downarrow$}&\textbf{IDs$\downarrow$}&\textbf{FM$\downarrow$}&\textbf{MOTA$\uparrow$}&\textbf{MOTP$\uparrow$}\\
\midrule
Iqbal \etal\cite{iqbal2017posetrack}& 63.0 & 64.8 & 775 & \textbf{502} & 431 & 5629 & 28.2 & 55.7\\
Ours & \textbf{65.9} & \textbf{83.2} &\textbf{949} &623&\textbf{202}&\textbf{3358}&\textbf{53.6}&\textbf{56.4}\\
\bottomrule
\end{tabular}}
\label{Table:PoseTrack Dataset}
\vspace{4mm}
\caption{Multi-person pose tracking results on PoseTrack dataset}
\end{table*}
\vspace{-2mm}

\begin{figure*}[!ht]
\centering
\includegraphics[scale=0.43]{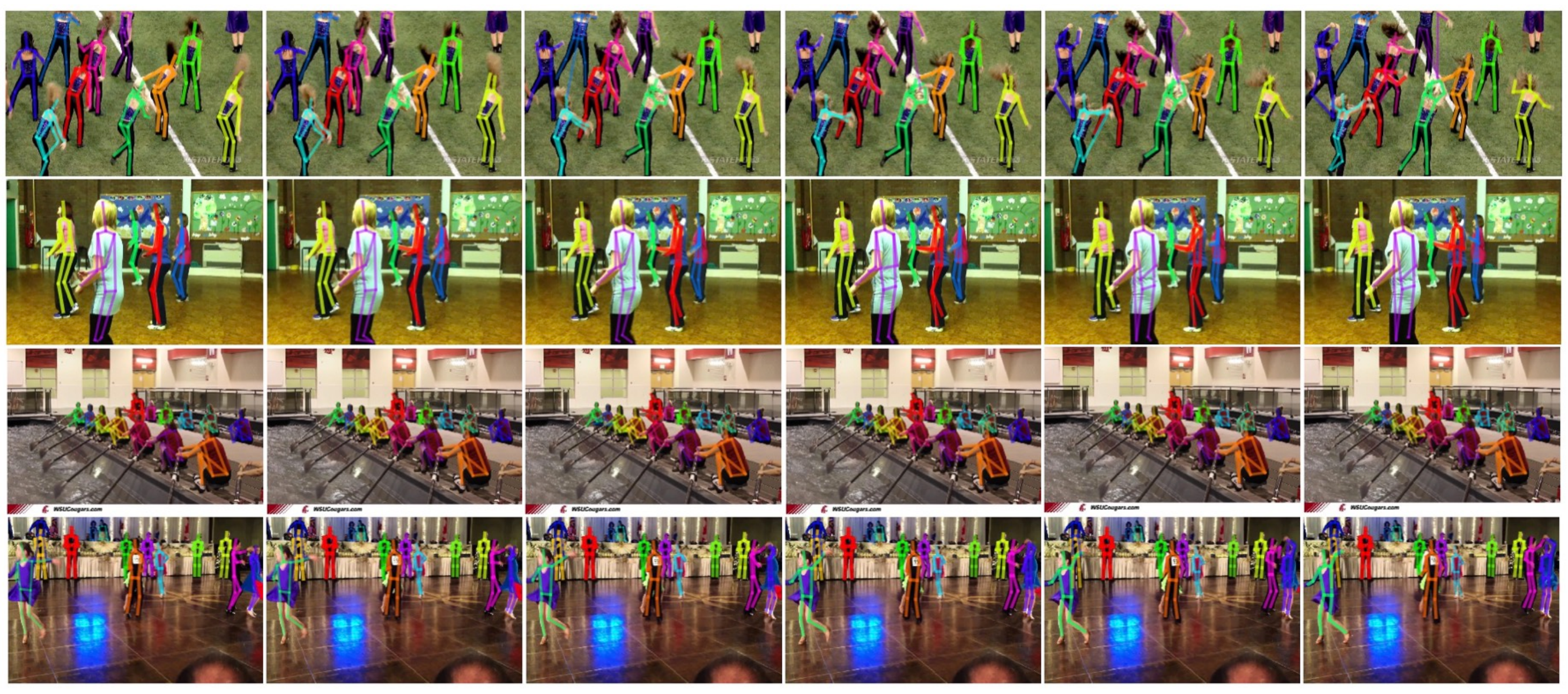} 
\caption{Some final posetracking results in videos}
\label{fig:result}
\vspace{-2mm}
\end{figure*}

PoseTrack Challenge Dataset is released in \cite{andriluka2017posetrack}. Selected and annotated like PoseTrack Dataset, it contains more videos. The testing dataset evaluation includes three tasks, but we only join Task2-Multi-Frame Person Pose Estimation, evaluated by mean average precision (mAP) and Task3-Pose tracking, evaluated by multi-object tracking accuracy (MOTA) metric. Tracking results of validation set and the test set of PoseTrack Challenge Dataset are presented in Table \ref{Table:MOTA}. Our method can achieve state-of-the-art results on validation and comparable results on test set. Some qualitative results are shown in Figure \ref{fig:result}.
\vspace{-2mm}
\begin{table*}[!ht]
\centering
\scalebox{0.61}{
\begin{tabular}{ccccccccc>{\columncolor[gray]{.8}}cccc}
\toprule
\textbf{Method}&\textbf{Dataset}&\textbf{\makecell{MOTA \\ Head}}&\textbf{\makecell{MOTA \\ Shou}}&\textbf{\makecell{MOTA \\ Elb}}&\textbf{\makecell{MOTA \\ Wri}}&\textbf{\makecell{MOTA \\ Hip}}&\textbf{\makecell{MOTA \\ Knee}}&\textbf{\makecell{MOTA \\ Ankl}}&\textbf{\makecell{MOTA\\Total}}&\textbf{\makecell{MOTP\\Total}}&\textbf{\makecell{Prcn}}&\textbf{\makecell{Rcll}}\\
\midrule
 Girdhar \etal \cite{girdhar2017detect} &\multirow{2}{*}{validation}& 61.7 & 65.5 & 57.3 & 45.7 & 54.3 & 53.1 & 45.7 & 55.2 & 61.5 & 88.1 & 66.5\\
Ours & ~ & 59.8 & 67.0 & 59.8 & 51.6 & 60.0 & 58.4 & 50.5 & \textbf{58.3} & 67.8 & 87.0 & 70.3 \\
\midrule
 Girdhar \etal\cite{girdhar2017detect} &{*(Mini)Test v1.0}& 55.9&59.0&51.9&43.9&47.2&46.3&40.1&49.6&34.1&81.9&67.4\\
Ours & {testset} & 52.0 & 57.4 & 52.8 & 46.6 & 51.0 & 51.2 & 45.3 & \textbf{51.0} & 16.9 & 78.9 & 71.2\\
\bottomrule
\end{tabular}}
\caption{Multi-person pose tracking results on PoseTrack Challenge dataset, * Note that this result was computed by online server on a subset of testset, and 51.8 MOTA is Girdhar \etal \cite{girdhar2017detect} got on full testset.}
\label{Table:MOTA}
\end{table*}
\vspace{-4mm}
\begin{table*}[!ht]
\centering
\scalebox{0.72}{
\begin{tabular}{ccccccccc>{\columncolor[gray]{.8}}c}
\toprule
\textbf{Method}&\textbf{Dataset}&\textbf{\makecell{Head \\ mAP}}&\textbf{\makecell{Shoulder \\ mAP}}&\textbf{\makecell{Elbow \\ mAP}}&\textbf{\makecell{Wrist \\ mAP}}&\textbf{\makecell{Hip \\ mAP}}&\textbf{\makecell{Knee \\ mAP}}&\textbf{\makecell{Ankle \\ mAP}}&\textbf{\makecell{Total \\ mAP}}\\
\midrule
 Iqbal \etal \cite{iqbal2017posetrack} & \multirow{2}{*}{PoseTrack} & 56.5 & 51.6 & 42.3 & 31.4 & 22.0 & 31.9 & 31.6 & 38.2\\
Ours & ~ & 64.7 & 65.9 & 54.8 & 48.9 & 33.3 & 43.5 & 50.6 & \textbf{51.7}\\
\midrule
 Girdhar \etal\cite{girdhar2017detect} & \multirow{2}{*}{PoseTrack Challenge(valid)}& 67.5 & 70.2 & 62 & 51.7 & 60.7 & 58.7 & 49.8 & 60.6
\\
Ours & ~ & 66.7 & 73.3 & 68.3 & 61.1 & 67.5 & 67.0 & 61.3 & \textbf{66.5} \\
\midrule
 Girdhar \etal\cite{girdhar2017detect} & {*(Mini)Test v1.0}
  & 65.3 & 66.7 & 59.7 & 51.2 & 58.6 & 55.8 & 48.8 & 58.5\\
Ours & {PoseTrack Challenge(test)}& 64.9 & 67.5 & 65.0 & 59.0 & 62.5 & 62.8 & 57.9 & \textbf{63.0} \\
\bottomrule
\end{tabular}}
\vspace{4mm}
\caption{Multi-person pose estimation results on all PoseTrack dataset,* Note that this result was computed by online server on a subset of test set, 59.6 mAP is Girdhar \etal \cite{girdhar2017detect} got on full testset.}
\label{Table:mAP}
\end{table*}
\vspace{-2mm}
\paragraph{Time Performance} Our proposed pose tracker is based on resulting poses in individual frames. That is, it is ready to apply in different multi-person pose estimators. The extra computation by our pose tracker is very minor, requiring 100ms per frame only. Therefore, it will not be the bottleneck of whole system, comparing to the speed of pose estimation.
\vspace{-3mm}   
\subsection{Training and Testing Details}
In this paper, we use ResNet152 based Faster R-CNN as human detector. Due to the absence of human proposal annotations, we generate human boxes by extending human keypoints boundary $20\%$ along both height and width directions, which are used for fine-tuning human detector. In the phrase of single person pose estimation training, we employed online hard example mining (OHEM) to deal with hard keypoints like hips and ankles. For each iteration, instead of sampling the highest $B/N$ losses in mini-batch, $k$ highest loss hard examples are selected. After selection, the SPPE update weights only from hard keypoints. These procedures increase slight computation time, but notably improve estimation performance of hips and ankles.

\subsection{Ablation Studies}
We evaluate the effectiveness of four proposed components: Deepmatching based Motion-guided box propagation (DMGP), improved deep proposal generator (iDPG), Pose Flow Builder (PF-Builder) and Pose Flow NMS (PF-NMS). The ablative studies are conducted on the validation of PoseTrack Challenge dataset, by removing these modules from the pipeline or replacing them with naive solvers, i.e., we replace the PF based tracker with box IoU based maximum weight bipartite matching tracker (IoU-tracker) used by \cite{girdhar2017detect}.

\vspace{-2mm}
\begin{table}[!ht]
\centering
\scalebox{.80}{
\begin{tabular}{l>{\columncolor[gray]{.8}}c>{\columncolor[gray]{.8}}cccc}
\toprule
\textbf{Method} &\textbf{\makecell{mAP}}&\textbf{\makecell{MOTA}}&\textbf{\makecell{MOTP}}&\textbf{\makecell{Prcn}}&\textbf{\makecell{Rcll}}\\
\midrule
\textbf{PoseFlow, full} &\textbf{66.5} &\textbf{58.3} & 67.8 & 87.0 & 70.3\\
\midrule
w/o PF-NMS &64.6 & 55.8&66.0 &82.2 &90.3\\
IoU-Tracker &64.6& 52.1& 61.2& 82.2&90.3\\
\midrule
w/o DMGP &62.2 & 53.7&63.4 &89.2 &62.3\\
w/o iDPG &65.4 & 57.8&66.9 &87.0 &70.3\\
\bottomrule
\end{tabular}}
\vspace{4mm}
\caption{Ablation comparison. ``IoU-Tracker'' means naive box IoU based matching tracker used by \cite{girdhar2017detect}. ``w/o PF-NMS'' means only using PF-Builder without PF-NMS. ``w/o DMGP'' means removing motion-guided box propagation. ``w/o iDPG'' means without improved deep proposal generator.}
\label{Ablation}
\end{table}
\vspace{-2mm}
\paragraph{PF-Builder and PF-NMS} PF-Builder is responsible for constructing pose flow. Due to its a global optimum solution, like Table \ref{Ablation} shows, it can guarantee better tracking performance than IoU-Tracker even without PF-NMS. PF-NMS can robustly merge redundant pose flows and re-link temporal disjoint ones, thus it can simultaneously polish pose estimation and tracking results by 1.9 mAP and 2.5 MOTA.
\vspace{-3mm}
\paragraph{DMGP and iDPG} DMGP is used for propagating adjacent boxes bidirectionally to recover missing boxes, so this module can improve tracking performance 4.6 MOTA by decreasing IDs dramatically. Because high recall of detections can fully exploit the power of PoseNMS module in RMPE framework \cite{fang2017rmpe}, 4.3 mAP is also increased thanks to this high recall. iDPG aims mainly to locate hard keypoints more accurately, because pose information is also leveraged during tracking, iDPG ultimately improve results by 1.1 mAP and 0.5 MOTA. 
\vspace{-1mm}

\section{Conclusion}
\vspace{-3mm}
We have presented a scalable and efficient top-down pose tracker, which mainly leverages spatio-temporal information to build pose flow to significantly boost pose tracking task. Two novel techniques, Pose Flow builder and Pose Flow NMS were proposed in this paper. In ablation studies, we prove that the combination of PF-Builder, PF-NMS, iDPG, and DMGP can guarantee a remarkable improvement in pose tracking tasks. Moreover, our proposed pose tracker that can process frames in a video at 10 FPS (excluding pose estimation in frames) has great potential in realistic applications. In the future, we would like to analyze long-term action recognition and scene understanding based the proposed pose tracker.

\clearpage
\newpage

\bibliography{egbib}

\end{document}